\documentclass{article}
\usepackage[utf8]{inputenc} % allow utf-8 input
\usepackage[T1]{fontenc}    % use 8-bit T1 fonts
\usepackage{microtype}      % microtypography

\usepackage{mathpazo}  % Modern font
\usepackage{helvet}    % Modern sans-serif font

\usepackage{amsmath}
\usepackage{amsfonts}
\usepackage{amssymb}
\usepackage{nicefrac}       % compact symbols for 1/2, etc.

\usepackage{graphicx}
\graphicspath{ {./images/} }
\usepackage{pgfplots}
\pgfplotsset{compat=1.18}
\usepgfplotslibrary{groupplots}
\usepackage{tikz}
\usepackage{svg}

\usepackage{booktabs}       % professional-quality tables
\usepackage{subcaption}     % For side-by-side figures

\usepackage{hyperref}       % hyperlinks
\usepackage{url}            % simple URL typesetting

\usepackage{xcolor}

\usepackage{lipsum}         % dummy text
\usepackage{filecontents}
\usepackage{scalerel}       % Added for \msquare command

\usepackage{arxiv}

\usepackage{graphicx}
\usepackage{pgfplots}
\usepackage{tikz}
\pgfplotsset{compat=1.18}

\usepackage{subcaption}

\usepackage{xcolor}

\usepackage{float}

\definecolor{model1}{RGB}{66,133,244}  % Google Blue
\definecolor{model2}{RGB}{234,67,53}   % Google Red
\definecolor{model3}{RGB}{251,188,4}   % Google Yellow
\definecolor{model4}{RGB}{52,168,83}   % Google Green
\definecolor{model5}{RGB}{103,58,183}  % Material Purple

\definecolor{bgColor}{RGB}{250,250,250}
\definecolor{gridColor}{RGB}{240,240,240}

\pgfplotscreateplotcyclelist{custom}{
    {model1, thick, mark=none},
    {model2, thick, mark=none},
    {model3, thick, mark=none},
    {model4, thick, mark=none},
    {model5, thick, mark=none}
}

\pgfplotsset{
    every axis/.append style={
        line width=0.5pt,
        tick style={thin, black!70},
        tick label style={font=\small},
        label style={font=\small},
        title style={font=\small\sffamily, align=center},
        legend style={font=\small\sffamily},
        grid=major,
        grid style={gridColor, line width=0.5pt},
        axis background/.style={fill=bgColor},
    }
}

\title{Dynamic Graph CNN with Jacobi Kolmogorov-Arnold Networks for 3D Classification of Point Sets}

\author{
Hanaa El Afia \\
ENSIAS \\
  Mohamed V University in Rabat\\
Rabat,Morocco  \\
  \texttt{hanaa\_elafia@um5.ac.ma} \\
   \And
Said Ohamouddou\\
ENSIAS \\
Mohamed V University in Rabat\\
Rabat, Morocco\\\texttt{said\_ohamouddou@um5.ac.ma} \\
  \And
Raddouane Chiheb \\
ENSIAS \\
Mohamed V University in Rabat\\
Rabat, Morocco\\\texttt{raddouane.chiheb@ensias.um5.ac.ma} \\
 \And
Abdellatif El Afia\\
ENSIAS \\
Mohamed V University in Rabat\\
Rabat, Morocco\\\texttt{abdellatif.elafia@ensias.um5.ac.ma} \\ 
}

\begin{document}
\maketitle
\begin{abstract}

We introduce Jacobi-KAN-DGCNN, a framework that integrates Dynamic Graph Convolutional Neural Network (DGCNN) with Jacobi Kolmogorov-Arnold Networks (KAN) for the classification of three-dimensional point clouds. This method replaces Multi-Layer Perceptron (MLP) layers with adaptable univariate polynomial expansions within a streamlined DGCNN architecture, circumventing deep levels for both MLP and KAN to facilitate a layer-by-layer comparison. In comparative experiments on the ModelNet40 dataset, KAN layers employing Jacobi polynomials outperform the traditional linear layer-based DGCNN baseline in terms of accuracy and convergence speed, while maintaining parameter efficiency. Our results demonstrate that higher polynomial degrees do not automatically improve performance, highlighting the need for further theoretical and empirical investigation to fully understand the interactions between polynomial bases, degrees, and the mechanisms of graph-based learning.

\end{abstract}

% keywords can be removed
%\keywords{First keyword \and Second keyword \and More}

\section{Introduction}

In recent years, there has been a significant surge in research interest shifting from two-dimensional (2D) images to three-dimensional (3D) data representations such as 3D point cloud, meshes, voxels, and depth images\cite{liu2024deep,wang2024vopifnet,miao2023occdepth}. This transition is driven by the growing demand for more detailed and accurate representations of real-world environments, which 2D images cannot fully capture. Among various 3D data formats, point clouds have emerged as a prominent representation due to their simplicity and efficiency in modeling complex geometries~\cite{liu20213d}.

3D point clouds offer several advantages, including the ability to represent objects without the constraints of a structured grid, making them ideal for capturing irregular and detailed surfaces~\cite{liu20213d}. They are widely used in applications like autonomous driving, robotics, and virtual reality, where precise spatial information is crucial~\cite{cui2021deep,pomerleau2015review,wirth2019pointatme}. The acquisition of point cloud data has become more accessible with advancements in sensing technologies such as LiDAR and depth cameras~\cite{liu20213d}.

A fundamental challenge in utilizing 3D point clouds is the classification and segmentation of the data~\cite{cui2021deep}. Classification involves assigning a label to an entire point cloud, while segmentation entails labeling each point within the cloud. These tasks are essential for object recognition, scene understanding, and interaction in 3D environments. Traditional deep learning methods have adapted to these challenges by employing multi-view projections or converting point clouds into voxel grids to leverage convolutional neural networks (CNNs)~\cite{pang20163d,huang2016point}. However, these approaches often suffer from information loss and high computational costs~\cite{qi2017pointnet}.

PointNet~\cite{qi2017pointnet} revolutionized point cloud processing by directly consuming raw point clouds using symmetric functions and multilayer perceptrons (MLPs). Despite its groundbreaking approach, PointNet has limitations in capturing local geometric features due to its reliance on global feature extraction~\cite{wang2019dynamic}. To address this, the Dynamic Graph CNN (DGCNN)~\cite{wang2019dynamic} was introduced as an alternative, utilizing graph-based methods to capture local structures and relationships between points. DGCNN employs MLPs for feature extraction, but MLPs can lack interpretability and may not effectively model complex interactions~\cite{liu2024kankolmogorovarnoldnetworks}.

In recent times, Kolmogorov–Arnold Networks (KANs) have been introduced as interpretable alternatives to multilayer perceptrons (MLPs)~\cite{liu2024kankolmogorovarnoldnetworks}. KANs are founded on the Kolmogorov–Arnold representation theorem, which asserts that any multivariate continuous function can be expressed by summing and compositions of univariate continuous functions~\cite{liu2024kankolmogorovarnoldnetworks}. Whereas MLPs utilize static activation functions at their nodes (neurons), KANs implement adaptable activation functions at their edges (weights). This characteristic renders KANs potentially more interpretable and adept at modeling intricate nonlinear relationships.

Although KANs have demonstrated potential in areas such as PDE solving~\cite{wang2024kolmogorov}, image classification~\cite{bodner2024convolutionalkolmogorovarnoldnetworks}, and time series analysis~\cite{xu2024kolmogorovarnoldnetworkstimeseries}, their utilization in 3D deep learning tasks, especially in the processing of 3D point cloud data, is still predominantly unexamined. To our knowledge, there is scant research on the integration of KANs with point cloud data in systems like as PointNet, as referenced in ~\cite{kashefi2024pointnet}. This gap prompts our exploration of the possible advantages of integrating KAN into graph-based techniques for point cloud analysis.

This study presents Jacobi-KAN-DGCNN, which incorporates the Jacobi Kolmogorov-Arnold Network (KAN) into the Dynamic Graph Convolutional Neural Network (DGCNN) architecture. Our primary aim is to investigate the impact of several KAN formulations—namely, Jacobian-based Jacobi KAN~\cite{jacob}, Legendre KAN (KALnet) \cite{torchkan}, and Discrete Chebyshev KAN (GRAM KAN) \cite{gram}—on the efficacy of point cloud categorization tasks. This integration seeks to explore how polynomial expansions can enhance the network's ability to capture complex geometric and topological features of data inputs. For focused comparisons, we have optimized the network architecture by omitting deep layers for both KAN and MLP components, concentrating instead on direct comparisons between a KAN layer and a standard linear layer. The principal contributions of this study are delineated as follows:

\begin{itemize}
    \item We benchmark DGCNN with KAN (Jacobi-KAN-DGCNN) and evaluate its performance against traditional MLP-based methods.
    \item We conduct an evaluation of the hyperparameters of Jacobi-KAN-DGCNN, specifically the degree and type of polynomial used in constructing KAN.
    \item We assess the efficiency of Jacobi-KAN-DGCNN on benchmarks for 3D object classification.
    \item We demonstrate that Jacobi-KAN-DGCNN achieves competitive performance compared to existing models, validating the effectiveness of our approach.
\end{itemize}

\section{Kolmogorov-Arnold Network (KAN)}
\label{Sect2}

Motivated by the Kolmogorov-Arnold representation theorem \cite{kolmogorov1957representation}, the Kolmogorov-Arnold Network (KAN) was recently proposed by \cite{liu2024kankolmogorovarnoldnetworks} as an innovative neural architecture. This theorem posits that each multivariate continuous function can be represented as a finite composition of continuous univariate functions, together with summations of intermediate outputs. Building upon this concept, polynomial-based variations of KAN, such as those employing discrete Chebyshev or Jacobi polynomials, have been investigated to improve approximation abilities, especially for discretized inputs like point clouds, pictures, and textual data.

Initially, we outline the fundamental construction of the single-layer KAN. We further delineate two polynomial bases—Jacobi and discrete Chebyshev—and demonstrate the integration of each into the KAN framework.

\subsection{General KAN Architecture}

Consider a \emph{single-layer} KAN that takes an input vector \(\mathbf{r}\in\mathbb{R}^{d_\text{input}}\) and outputs a vector \(\mathbf{s}\in\mathbb{R}^{d_\text{output}}\). The mapping is given by
\begin{equation}
\mathbf{s}_{d_\text{output}} 
\;=\; 
\mathcal{A}_{d_\text{output}\times d_\text{input}} \,\mathbf{r}_{d_\text{input}},
\tag{1}
\label{Eq1}
\end{equation}
where \(\mathcal{A}_{d_\text{output}\times d_\text{input}}\) is a learnable tensor of univariate functions:
\begin{equation}
\mathcal{A}_{d_\text{output}\times d_\text{input}}
=
\begin{bmatrix}
\psi_{1,1}(\cdot) & \psi_{1,2}(\cdot) & \cdots & \psi_{1,d_\text{input}}(\cdot) \\
\psi_{2,1}(\cdot) & \psi_{2,2}(\cdot) & \cdots & \psi_{2,d_\text{input}}(\cdot) \\
\vdots            & \vdots            & \ddots & \vdots                        \\
\psi_{d_\text{output},1}(\cdot) & \psi_{d_\text{output},2}(\cdot) & \cdots & \psi_{d_\text{output},d_\text{input}}(\cdot)
\end{bmatrix}.
\tag{2}
\label{Eq2}
\end{equation}

Each \(\psi_{i,j}(\gamma)\) is a polynomial expansion (or another univariate functional form) evaluated at the corresponding input component \(\gamma \in \mathbf{r}\). 

\subsection{Jacobi Polynomial--Based KAN}
\label{jacobi-sec}

\subsubsection{Polynomial Definition}

In this variant, each univariate function \(\psi(\gamma)\) expands as a linear combination of Jacobi polynomials \(f_i^{(\alpha,\beta)}(\gamma)\):
\begin{equation}
\psi(\gamma) \;=\; \sum_{i=0}^{n} \omega_i \; f_i^{(\alpha,\beta)}(\gamma),
\tag{3}
\label{Eq3Jacobi}
\end{equation}
where:
\begin{itemize}
    \item \(n\) is the polynomial order,
    \item \(\{\omega_i\}\) are trainable parameters,
    \item \(f_i^{(\alpha,\beta)}(\gamma)\) is the Jacobi polynomial of degree \(i\) with parameters \(\alpha\) and \(\beta\).
\end{itemize}

\subsubsection{Recursive Computation of Jacobi Polynomials}

Jacobi polynomials \(f_n^{(\alpha,\beta)}(\gamma)\) can be computed via the three-term recurrence:
\begin{equation}
f_{n}^{(\alpha,\beta)}(\gamma) 
\;=\; 
\bigl(a_n \,\gamma + b_n\bigr)\,f_{n-1}^{(\alpha,\beta)}(\gamma) 
\;+\; 
c_n \,f_{n-2}^{(\alpha,\beta)}(\gamma),
\tag{4}
\label{Eq4Jacobi}
\end{equation}
where the coefficients are
\begin{align}
a_n &= \frac{(2n+\alpha+\beta-1)\,(2n+\alpha+\beta)}{2n\,(n+\alpha+\beta)}, \notag\\
b_n &= \frac{(2n+\alpha+\beta-1)\,(\alpha^2 - \beta^2)}{2n\,(n+\alpha+\beta)\,(2n+\alpha+\beta-2)}, \notag\\
c_n &= \frac{-2\,(n+\alpha-1)\,(n+\beta-1)\,(2n+\alpha+\beta)}{2n\,(n+\alpha+\beta)\,(2n+\alpha+\beta-2)}.
\tag{5--7}
\label{Eq567Jacobi}
\end{align}

The initial conditions are:
\begin{equation}
f_0^{(\alpha,\beta)}(\gamma) = 1,
\quad
f_1^{(\alpha,\beta)}(\gamma) = \frac{1}{2}(\alpha + \beta + 2)\,\gamma \;+\; \frac{1}{2}(\alpha - \beta).
\tag{8--9}
\label{Eq89Jacobi}
\end{equation}

\subsubsection{Implementation Notes}

\paragraph{Input Scaling:}
Because Jacobi polynomials are typically defined on \([-1, 1]\), each input component \(r_j\in \mathbf{r}\) is first mapped to \([-1, 1]\), often via a \(\tanh\) function.

\paragraph{Recursive Computation:}
Jacobi polynomials are computed in ascending order (\(0\) to \(n\)) using the recurrence relation.

\paragraph{Special Cases:}
The following parameter choices recover well-known polynomial families:

\begin{itemize}
    \item \textbf{Legendre polynomials}: Parameters \(\alpha=0, \beta=0\).
    \item \textbf{Chebyshev polynomials of the first kind}: Set \(\alpha=\beta=-0.5\).
    \item \textbf{Chebyshev polynomials of the second kind}: Set \(\alpha=\beta=0.5\).
    \item \textbf{Discrete Chebyshev polynomials (Discrete Chebyshev KAN)}: These do not utilize the \(\alpha\) and \(\beta\) parameters as in their continuous counterparts but are instead a distinct group typically employed for \textit{discretized inputs}. For further information, consult Section~\ref{discrete-cheb-sec}.
\end{itemize}

\subsection{Discrete Chebyshev Polynomial--Based KAN}
\label{discrete-cheb-sec}

In the discrete Chebyshev variant of KAN, each univariate function \(\psi(\gamma)\) is expanded in terms of discrete Chebyshev polynomials \(P_i(\gamma)\). This section outlines their definition and usage.

\subsubsection{Discrete Chebyshev Polynomial Definition}

\begin{equation}
\psi(\gamma) = \sum_{i=0}^n \beta_i \; P_i(\gamma),
\tag{10}
\label{Eq10Cheb}
\end{equation}
where:
\begin{itemize}
    \item \(n\) is the polynomial degree,
    \item \(\{\beta_i\}\) are trainable parameters,
    \item \(P_i(\gamma)\) are the discrete Chebyshev polynomials of degree \(i\).
\end{itemize}

\subsubsection{Recursive Construction}

Define:
\begin{equation}
P_0(\gamma) = 1, 
\quad
P_1(\gamma) = \gamma.
\tag{11--12}
\label{Eq1112Cheb}
\end{equation}
For \(n \ge 2\), use the recurrence:
\begin{equation}
P_{n}(\gamma) 
= 
2\,\gamma \, P_{n-1}(\gamma) 
- 
\beta(n-1,n)\,P_{n-2}(\gamma),
\tag{13}
\label{Eq13Cheb}
\end{equation}
where \(\beta(n-1,n)\) is a weight adjustment factor for numerical stability:
\begin{equation}
\beta(n-1,n) 
= 
\frac{\bigl(n + (n-1)\bigr)\,\bigl(n - (n-1)\bigr)}{4n^2 - 1} 
\;\beta_{n-1}.
\tag{14}
\label{Eq14Cheb}
\end{equation}

\paragraph{Note:} The exact form of \(\beta(n-1,n)\) can vary by implementation and may be tuned for stability in specific tasks.

Because each \(\psi(\gamma)\) has \((n+1)\) learnable coefficients \(\omega_i\), the total number of trainable parameters in \(\mathcal{A}\) for a single KAN layer is \((n+1)\times d_\text{input} \times d_\text{output}\).

\section{Overview of DGCNN and Its Key Principles}

The proposed \textit{Dynamic Graph Convolutional Neural Network (DGCNN)} introduces a novel framework for learning on 3D point clouds by leveraging dynamic graph-based feature aggregation. Unlike static graph-based models, DGCNN dynamically updates the graph structure at each layer, enabling adaptive feature learning. The model integrates local and global geometric structures using \textit{Edge Convolution (EdgeConv)} operations and achieves efficient non-local information propagation.

\subsection{Key Components of DGCNN}

\paragraph*{1. Dynamic Graph Construction.}
Given an $F$-dimensional point cloud $\mathbf{X} = \{\mathbf{x}_1, \ldots, \mathbf{x}_n\} \subseteq \mathbb{R}^F$, DGCNN constructs a dynamic $k$-nearest neighbor (k-NN) graph $\mathcal{G}^{(l)}$ for each layer $l$, where $\mathcal{G}^{(l)} = (\mathcal{V}, \mathcal{E}^{(l)})$. Graph edges are updated dynamically based on proximity in the learned feature space:
\[
\mathcal{E}^{(l)} = \{(i, j): j \in \text{kNN}(\mathbf{x}^{(l)}_i, k)\},
\]
where $\mathbf{x}^{(l)}_i$ is the feature of the $i$-th point at layer $l$.

\paragraph*{2. Edge Convolution (EdgeConv).}
EdgeConv captures both local and global geometric features by defining edge features as:
\[
\boldsymbol{e}_{ij} = h_{\boldsymbol{\Theta}}(\mathbf{x}_i, \mathbf{x}_j),
\]
where $h_{\boldsymbol{\Theta}}: \mathbb{R}^F \times \mathbb{R}^F \to \mathbb{R}^{F'}$ is a learnable function. The aggregated feature at each vertex $i$ is computed using a symmetric operation $\square$ (e.g., $\max$ or $\sum$):
\[
\mathbf{x}'_i = \mathop{\square}_{j: (i,j) \in \mathcal{E}} h_{\boldsymbol{\Theta}}(\mathbf{x}_i, \mathbf{x}_j).
\]
This mechanism enables permutation invariance and partial translation invariance.

\paragraph*{3. Adaptive Feature Learning.}
By recomputing the graph at each layer, DGCNN ensures non-local information diffusion and learns the optimal graph structure dynamically. This process allows for capturing hierarchical semantic relationships within the point cloud.

\paragraph*{4. Integration of Local and Global Features.}
The edge function $h_{\boldsymbol{\Theta}}(\mathbf{x}_i, \mathbf{x}_j)$ combines local structure, represented by $\mathbf{x}_j - \mathbf{x}_i$, and global context, represented by $\mathbf{x}_i$. For instance, a shared MLP is used to implement:
\[
e_{ijm}' = \mathrm{ReLU}(\boldsymbol{\theta}_m \cdot (\mathbf{x}_j - \mathbf{x}_i) + \boldsymbol{\phi}_m \cdot \mathbf{x}_i),
\]
with final aggregation as:
\[
x'_{im} = \max_{j: (i,j) \in \mathcal{E}} e_{ijm}'.
\]

\subsection{Comparison to Existing Methods}
DGCNN unifies principles from PointNet and graph CNNs. Unlike PointNet, which ignores local geometric structure, and traditional graph CNNs, which rely on fixed graphs, DGCNN dynamically constructs and updates graphs during training, enabling effective local-global feature integration. As shown in Table~\ref{table:sum}, DGCNN is versatile and achieves state-of-the-art performance by learning adaptive graph structures and feature representations.

\subsection{Properties of DGCNN}
DGCNN possesses key properties critical for point cloud analysis:
\begin{itemize}
    \item \textbf{Permutation Invariance:} Achieved via symmetric aggregation functions (e.g., $\max$).
    \item \textbf{Translation Invariance:} Enabled by using relative coordinates $\mathbf{x}_j - \mathbf{x}_i$.
    \item \textbf{Scalability:} Supports large point clouds with efficient dynamic graph updates.
\end{itemize}

DGCNN demonstrates that learning both local and global features on dynamically constructed graphs is a powerful paradigm for 3D shape analysis.

\section{Architecture of Jacobi-KAN-DGCNN}
\label{sec:dgcnnkan}

The Jacobi-KAN-DGCNN architecture consists of a sequence of Dynamic Edge Convolutions (EdgeConv) followed by a series of KAN layers. Each EdgeConv operation is paired with a KAN layer, which processes the edges dynamically constructed at each layer of the network. The design is crafted to capture both local and global features effectively, using the recursive nature of the Chebyshev polynomials to handle the data efficiently.

Following the initial feature transformation through the EdgeConv and KAN layers, the network utilizes global pooling operations to aggregate the features, which are then further processed through additional KAN layers to predict the final outputs. The architecture ensures permutation invariance and captures intricate geometric structures dynamically through its graph-based approach.

\begin{figure}[h!]
    \centering
    \includegraphics[width=16cm]{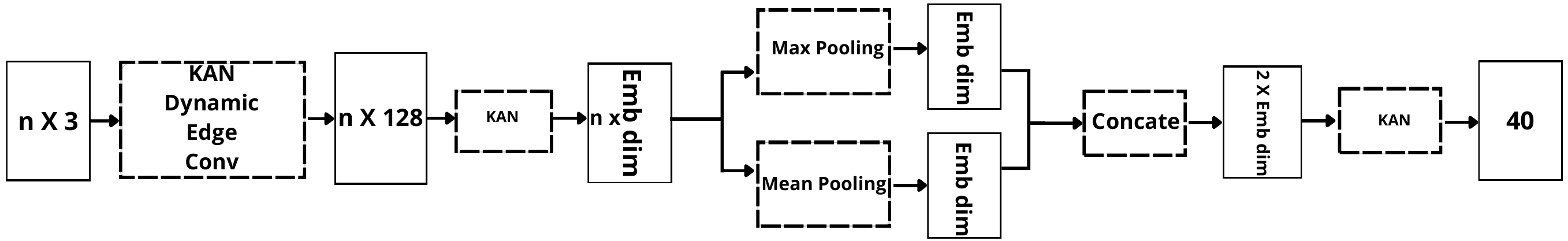} % Adjust width as needed
    \caption{The Jacobi-KAN-DGCNN architecture}
    \label{fig:model} % Optional: for referencing the figure
\end{figure}

For simplicity reasons and to facilitate comparison, we utilize a simplified DGCNN architecture consisting of:

A single EdgeConv cell, which processes the geometric features of the point cloud
A KAN layer that captures the contextual relationships between points
A pooling-based aggregation mechanism as depicted in Figure ~\ref{fig:model}, which combines:

Max pooling branch: capturing the most prominent features
Mean pooling branch: preserving average feature distributions

This streamlined architecture maintains the essential components of DGCNN while reducing complexity, allowing for more direct comparisons with other approaches. The concatenation of max and mean pooling features ensures we retain both local and global structural information from the point cloud data.

\section{Convergence}

The global convergence of the model relies on three pillars: local convergence, stable propagation, and loss stability
\subsection{local Convergence }
KAN layers use Chebyshev polynomials to approximate functions in the graph feature space. The approximation of a function  $f(x)$ by a Chebyshev series is given by:

\begin{equation}
f(x) \;\approx \sum_{i=0}^{n} \omega_i \; f_i^{(\alpha,\beta)}(x),
\tag{3}
\label{Eq3Jacobi}
\end{equation}
where
\begin{itemize}
    \item \(n\) denotes the order of the polynomial,
    \item \(\{\omega_i\}\) represent the trainable parameters,
    \item \(f_i^{(\alpha,\beta)}(\gamma)\) corresponds to the Jacobi polynomial of degree \(i\) with parameters \(\alpha\) and \(\beta\).
\end{itemize}

\textbf{Truncation error}:\\
The error between the actual function $f(x)$  and its approximation by the truncated series is:
\[
E_T = \left\| f(x) -  \sum_{i=0}^{n} \omega_i \; f_i^{(\alpha,\beta)}(x) \right\| \leq \max_{x \in [-1, 1]} \left| \omega_i f_i^{(\alpha,\beta)}(x) \right|.
\]

For an analytic function, the coefficients $\omega_i $ decrease exponentially:
\[
|\omega_i | \leq C \cdot \rho^{-n},
\]
where $C > 0$ and  $\rho > 1$ depend on the regularity of $f(x)$. This guarantees that the truncation error satisfies :
\[
E_T = \mathcal{O}(\rho^{-n}),
\]
with $N$ the number of terms in the series.

KAN layers apply this approximation to the local features of dynamic graphs. Due to the exponential decrease of the coefficients, each layer reduces the local error with precise control by increasing the number of terms in the series.$N$.

Each KAN layer effectively reduces $E_T$, contributing to local convergence.
\subsection{Stable propagation: Graph dynamics and global coherence}

\textbf{Update of dynamic graphs}

Dynamic graphs are reevaluated at each layer to reflect the updated local relationships. This aspect is central to models based on dynamic graphs, such as DGCNN. The update of the graphs is described by the following formula:
\[
A^{(l+1)} = f_{\text{EdgeConv}} \left( A^{(l)}, X^{(l)} \right),
\]
où :
\begin{itemize}
    \item $A^{(l)}$ The adjacency matrix of the graph at the layer.$l$.
    \item $X^{(l)}$ Are the node features at the layer. $l$.
\end{itemize}

The calculation of dynamic graphs is often based on the $k$-nearest neighbors in feature space $X^{(l)}$, which allows local connections to be dynamically adapted as characteristics change.

 \textbf{Graph Stability}

\begin{itemize}
    \item \textbf{Permutation Invariance}: 
    Permutation invariance means that the order of the nodes does not affect the computations. This property is essential because graphs do not depend on the indexing of the nodes. Operators like \textit{EdgeConv} ensure this invariance by constructing graphs solely from the local distances between node features, regardless of their order.
    \item \textbf{Local Contraction}: 
    Local contraction is a mathematical condition that guarantees dynamic graphs converge to a stable structure over iterations. It is described by the following formula:
\end{itemize}
\[
\| A^{(l+1)} - A^* \| \leq \alpha \| A^{(l)} - A^* \|,
\]
with :
\begin{itemize}
    \item $\alpha \in [0, 1[$, a constant.
    \item $A^*$, the optimal adjacency matrix.
\end{itemize}

When $\alpha < 1$, the distance between $A^{(l+1)}$ and $A^*$ decreases with each iteration. This ensures that graphs become increasingly coherent and converge on a stable structure after several layers. 

This behaviour is natural, since the transformations carried out by $f_{\text{EdgeConv}}$ (based on distances in feature space) tend to reduce errors between successive graphs.

\textbf{Consistent Propagation}:\\
The demonstration asserts the following points:
\begin{itemize}
    \item Local errors are reduced at each layer through the evolution of features and local connections.
    \item Dynamic graphs become stable, ensuring global consistency.
\end{itemize}
\subsection{Loss Stability: Global Optimization via Gradient Descent}
The global convergence of the model relies on the optimization of the loss function, which combines the minimization of local errors in each layer and the global loss.\\

\textbf{ General framework}\\
We consider a loss function $L(\theta)$, where $\theta$ groups the model parameters. The aim is to minimize $L(\theta)$ by applying gradient descent. The parameter update is given by :
\[
\theta^{(t+1)} = \theta^{(t)} - \eta \nabla_\theta L(\theta^{(t)}),
\]
where:
\begin{itemize}
    \item $\eta > 0$ is the learning rate,
    \item $\nabla_\theta L(\theta^{(t)})$ is the gradient of $L$ evaluated in $\theta^{(t)}$.
\end{itemize}

\textbf{Necessary assumptions}\\
To guarantee convergence, we make two assumptions:
 \begin{itemize}
     \item \textbf{Convexity}:\\ The function $L(\theta)$ is convex if, for all $\theta_1, \theta_2 \in \mathbb{R}^d$ and $\alpha \in [0,1]$, we have :
\[
L(\alpha \theta_1 + (1-\alpha) \theta_2) \leq \alpha L(\theta_1) + (1-\alpha) L(\theta_2).
\]
\item \textbf{ Gradient Lipschitz}:\\  The gradient of  $L(\theta)$ is Lipschitz-continuous with constant $L > 0$, which means that :
\[
\| \nabla_\theta L(\theta_1) - \nabla_\theta L(\theta_2) \| \leq L \| \theta_1 - \theta_2 \|.
\]
 \end{itemize}
\textbf{Decrease in function $L(\theta)$}\\
Using a second-order Taylor expansion around $\theta^{(t)}$, we can write :
\[
L(\theta) \approx L(\theta^{(t)}) + \nabla_\theta L(\theta^{(t)})^\top (\theta - \theta^{(t)}) + \frac{L}{2} \| \theta - \theta^{(t)} \|^2.
\]

By applying the update $\theta^{(t+1)} = \theta^{(t)} - \eta \nabla_\theta L(\theta^{(t)})$, we obtain :
\[
L(\theta^{(t+1)}) - L(\theta^{(t)}) \leq -\eta \| \nabla_\theta L(\theta^{(t)}) \|^2 + \frac{\eta^2 L}{2} \| \nabla_\theta L(\theta^{(t)}) \|^2.
\]
By grouping the terms :
\[
L(\theta^{(t+1)}) \leq L(\theta^{(t)}) - \left( \eta - \frac{\eta^2 L}{2} \right) \| \nabla_\theta L(\theta^{(t)}) \|^2.
\]

\textbf{ Condition on learning rate}\\

To guarantee a reduction in $L(\theta)$, the term $\eta - \frac{\eta^2 L}{2}$ must be positive. This imposes an upper bound on $\eta$ :
\[
\eta \leq \frac{2}{L}.
\]

With a $\eta$ respecting this condition, the loss function decreases with each iteration.

\textbf{Convergence to a Minimum}\\
By summing the variations $L(\theta^{(t+1)}) - L(\theta^{(t)})$ over $t$ iterations, and under the assumption $\|\nabla_\theta L(\theta)\| \neq 0$, it can be shown that $L(\theta)$ continues to decrease. Under the convexity assumption, it is guaranteed that:
\[
\theta^{(t)} \to \theta^*, \quad \text{where } \nabla_\theta L(\theta^*) = 0.
\]

\textbf{Rate of Convergence: $\mathcal{O}(1/t)$}\\

For a convex and differentiable function with a Lipschitz gradient, it is demonstrated that the difference between $L(\theta^{(t)})$ and the minimum $L^*$ decreases with the number of iterations:
\[
L(\theta^{(t)}) - L^* \leq \frac{C}{t},
\]
where $C$ is a constant depending on the initial conditions.

Thus, the loss decreases at a rate of $\mathcal{O}(1/t)$, reflecting the typical sublinear convergence of gradient descent algorithms.

\textbf{ Combined error minimization}\\
Global optimization combines the reduction of local errors in each layer (by KAN) and the minimization of global loss. The total loss function is defined by :
\[
L_{\text{total}} = L + \lambda \sum_l \| E_T^{(l)} \|,
\]
where $\lambda$ is a regularization coefficient.

This formulation guarantees that :
\[
L_{\text{total}} \to L^* \quad  t \to \infty.
\]

\textbf{ Conclusion}\\
Under assumptions of convexity and regularity, gradient descent guarantees convergence of the overall loss $L_{\text{total}}$ to a minimum $L^*$. This convergence is based on a balanced optimization between local errors and global regularization.
\section{Experimental Results and Analysis}
We evaluate our proposed Jacobi-based Jacobi-KAN-DGCNN model ($\alpha= \beta = 1$) on the ModelNet40 dataset \cite{modelnet}, a comprehensive 3D shape classification benchmark. The dataset comprises 12,311 models spanning 40 categories, with 9,843 models designated for training and 2,468 for testing. Our preprocessing pipeline includes uniform sampling of 1,024 points, followed by data augmentation through random scaling (2/3, 3/2) and random shifting (0.2), concluded by sphere normalization. We conducted a layer-by-layer comparison and therefore did not use deep architectures for both KAN and MLP.
\subsection{Evaluation Metrics}
We employ two primary metrics to assess model performance:
Overall Accuracy (OA) quantifies the global classification performance across all samples:
\begin{equation}
\text{OA} = \frac{\sum_{i=1}^{N} \delta(\hat{y}_i, y_i)}{N},
\end{equation}
where N represents the total sample count, $\hat{y}_i$ denotes the predicted label, $y_i$ represents the ground truth label, and $\delta(\hat{y}_i, y_i)$ is an indicator function returning 1 for correct predictions ($\hat{y}_i = y_i$) and 0 otherwise.
Mean Class Accuracy (MCA) provides a balanced perspective across all classes, mitigating the impact of class imbalance:
\begin{equation}
\text{MCA} = \frac{1}{C} \sum_{c=1}^{C} \frac{\sum_{i=1}^{N_c} \delta(\hat{y}_i, y_i)}{N_c},
\end{equation}
where C denotes the total number of classes and $N_c$ represents the sample count in class c.
\subsection{Implementation Details}
All experiments were conducted using PyTorch Geometric on dual RTX A4000 GPUs. Hyperparameter configurations are detailed in Table~\ref{tab:kan-params}.

\begin{table}[h]
\centering
\caption{Hyperparameters of Experiments}
\label{tab:kan-params}
\begin{tabular}{@{}ccccccccccc@{}}
\toprule
\textbf{Aggregator} & \textbf{Batch Size} & \textbf{Embedding Dimension} & \textbf{Epochs} & \textbf{K} & \textbf{Learning Rate} & \textbf{Momentum} & \textbf{Optimizer} &  \textbf{Points} \\
\midrule
max & 16 & 1024 & 250  & 5 & 0.001 & 0.9 & SGD & 1024 \\
\bottomrule
\end{tabular}
\end{table}

\subsection{Comparative Analysis}
\subsubsection{Model Architecture Comparison}
We first compare our Jacobi Jacobi-KAN-DGCNN against three baseline architectures: an MLP-based DGCNN, a Discrete Chebyshev-based approach (GRAM-KAN), and a Legendre Polynomials-based implementation (KAL-Net). Table~\ref{tab:model_accuracies} presents the comparative results.

\begin{table}[ht]
	\centering
	\caption{Comparison of Model Accuracies}
	\label{tab:model_accuracies}
	\begin{tabular}{llrrr}
		\toprule
		{} & Model & OC & MCA & Trainable Parameters \\
		\midrule
		0 & GRAM KAN  & 0.817666 & 0.812384 & 1,071,196 \\
		1 & \textbf{Jacobi KAN degree 3}   & \textbf{0.873177} & \textbf{0.838488} & 1,071,184 \\
		2 & KAL-Net    & 0.689222 & 0.669151 & 1,071,184 \\
		3 & MLP        & 0.844814 & 0.774360 & 214,952 \\
		\bottomrule
	\end{tabular}
\end{table}

The Jacobi KAN with degree 3 demonstrates superior performance, achieving an overall accuracy of 87.32\% and mean class accuracy of 83.85\%. This represents a significant improvement over both the traditional MLP approach (84.48\% OA) and other KAN variants. Notably, this performance is achieved with a comparable parameter count to GRAM-KAN, suggesting improved parameter efficiency.

\begin{figure}[htbp]
	
	\centering
	% Subfigure for Loss
    \includegraphics[width=\textwidth]{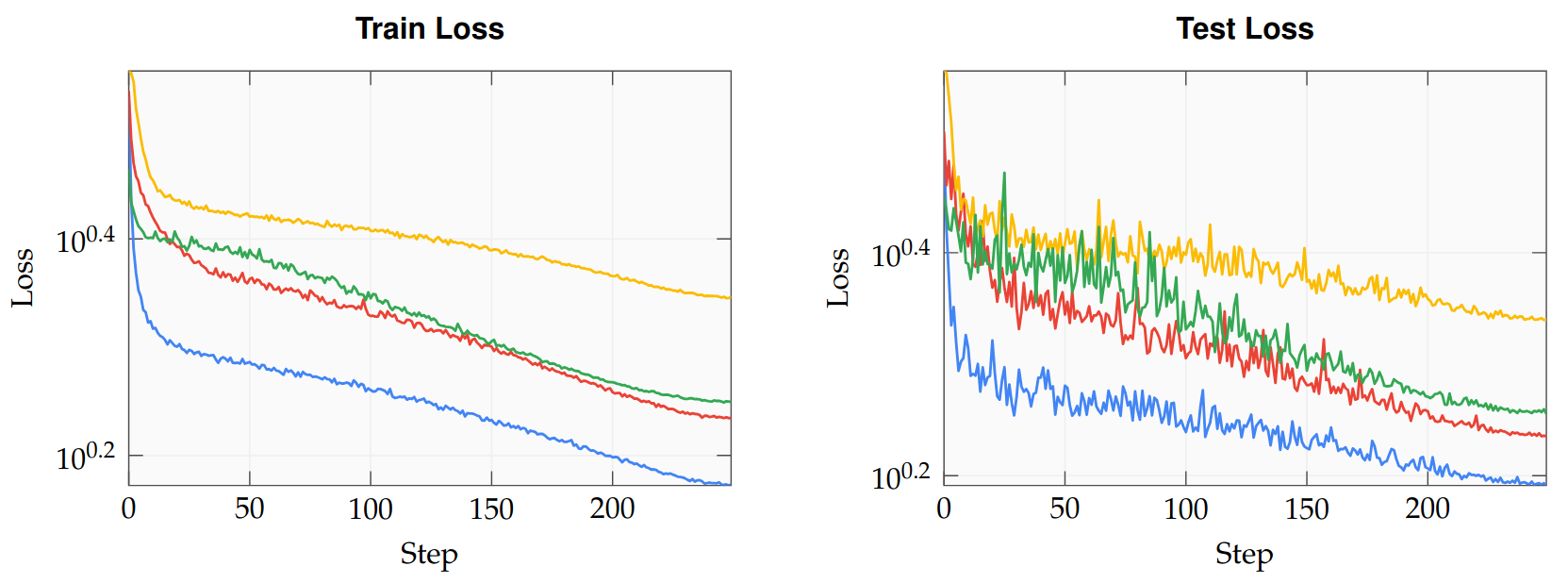}
	\caption{Comparison of Train and Test Loss across Different Models}
	\label{fig:loss1}
\end{figure}

\begin{figure}[htbp]
	\centering
	% Subfigure for Accuracy
    \includegraphics[width=\textwidth]{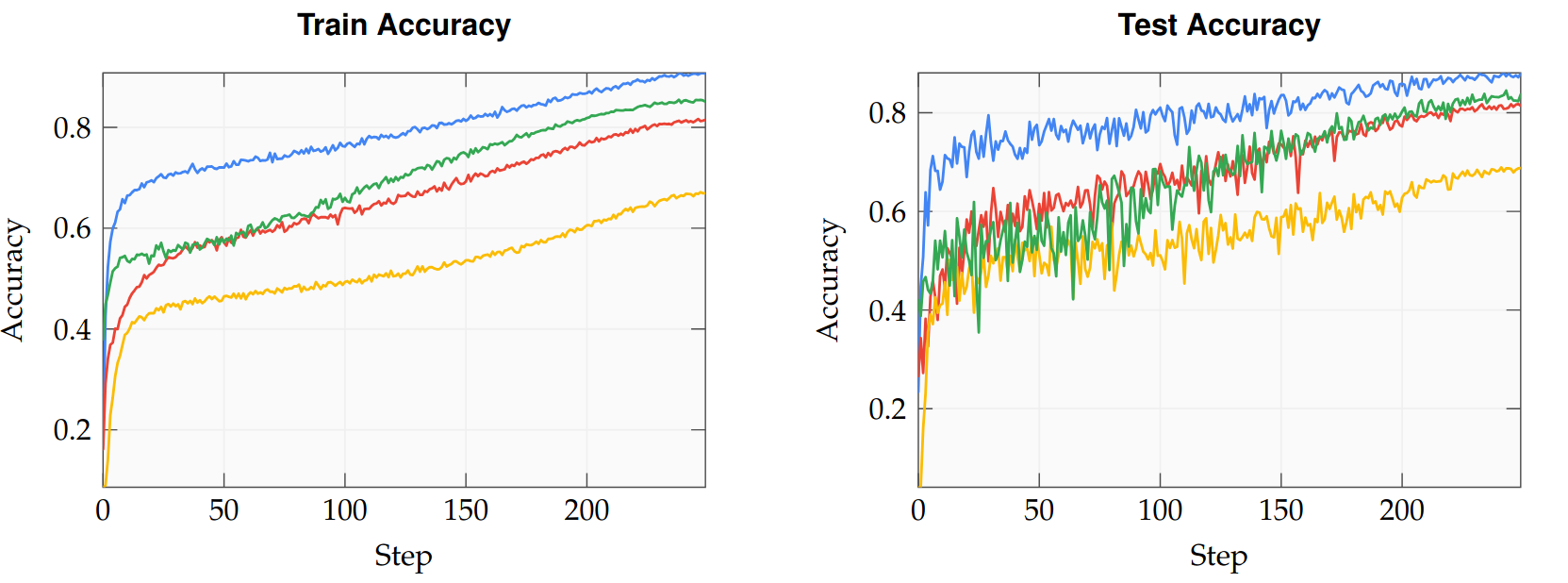}
	\caption{Comparison of Train and Test Accuracy across Different Models}
	\label{fig:acc1}
\end{figure}

\begin{figure}[htbp]
	
	\centering
	% Subfigure for avg_per_class_accuracy
    \includegraphics[width=\textwidth]{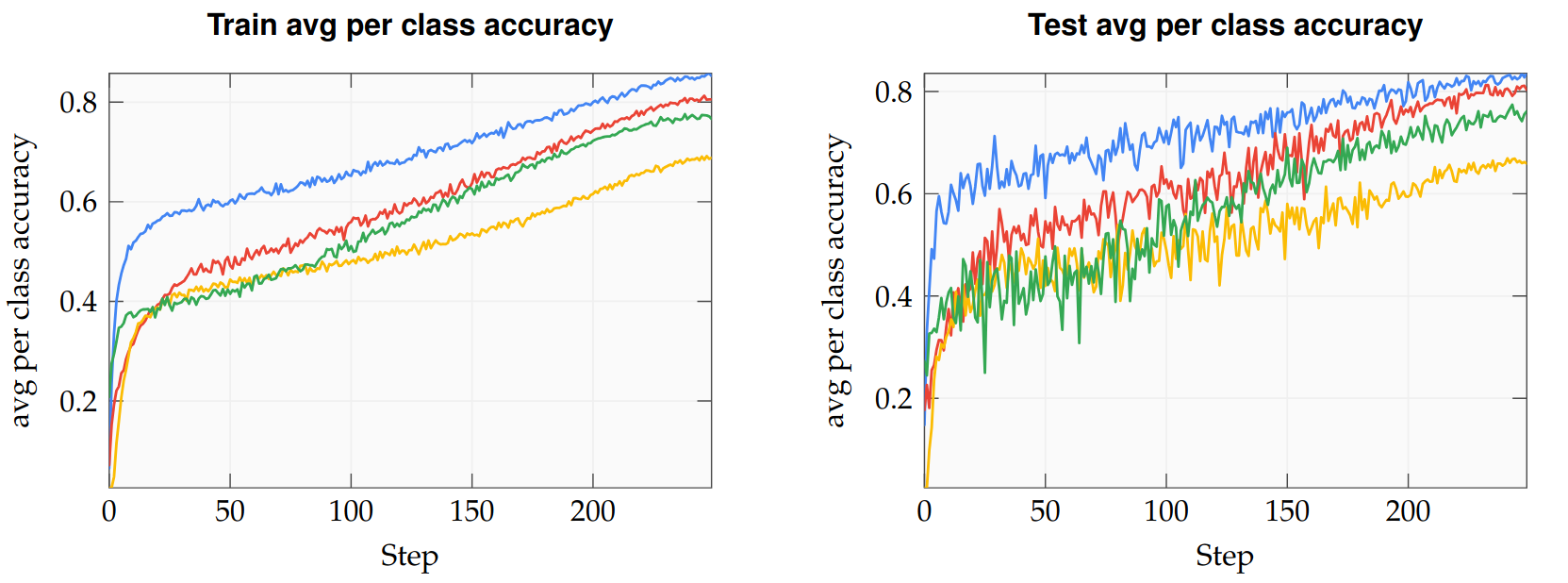}
	\caption{Comparison of Train and Test avg per class accuracy across Different Models}
	\label{fig:avg_acc1}
\end{figure}

The convergence analysis (Figures~\ref{fig:loss1},\ref{fig:acc1},\ref{fig:avg_acc1}) reveals that the Jacobi KAN exhibits faster and more stable convergence compared to baseline models. While both MLP and GRAM-KAN show comparable convergence patterns, with GRAM-KAN displaying marginal improvements, KAL-Net demonstrates notably slower convergence.
\subsubsection{Impact of Polynomial Degree}

To investigate the relationship between model complexity and performance, we conducted experiments varying the Jacobi polynomial degree (Table~\ref{tab:jacobikan_comparison}). Interestingly, the degree-1 model achieves the highest overall accuracy (87.68\%) while maintaining the lowest parameter count (643,664). This suggests that increased model complexity does not necessarily translate to improved performance.

\begin{table}[h]
	\centering
	\caption{Comparison of JacobiKAN Performance across Different Degrees}
	\label{tab:jacobikan_comparison}
	\begin{tabular}{lccc}
		\toprule
		Model & OA & MCA & Trainable Parameters \\
		\midrule
		\textbf{JacobiKAN degree 1 }& \textbf{0.876823} & 0.813733 & \textbf{643,664} \\
		JacobiKAN degree 2 & 0.859400 & 0.820047 & 857,424 \\
		JacobiKAN degree 3 & 0.873177 & 0.838488 & 1,071,184 \\
		JacobiKAN degree 4 & 0.880875 & 0.835413 & 1,284,944 \\
		\bottomrule
	\end{tabular}
\end{table}

\begin{figure}[htbp]
	
	\centering
	% Subfigure for lossuracy
    \includegraphics[width=\textwidth]{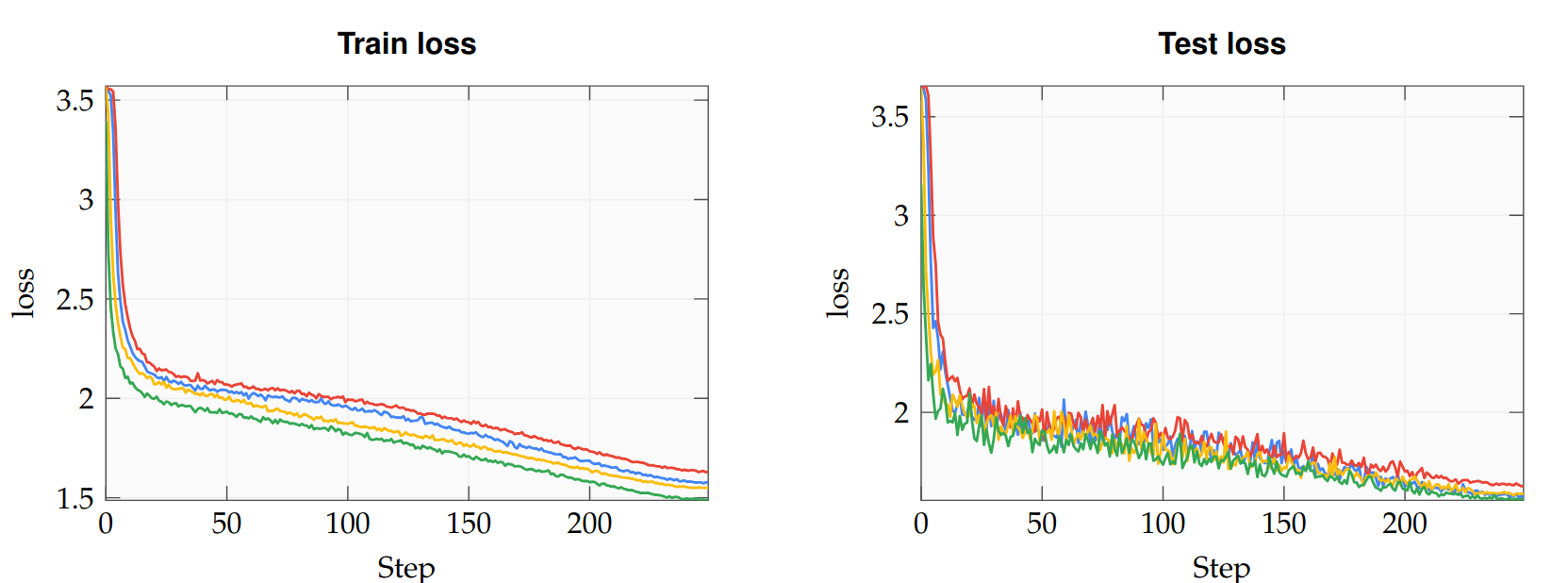}
	\caption{comparison of Train and Test lossuracy for Varying Degree Values}
	\label{fig:loss2}
\end{figure}

\begin{figure}[htbp]
	\centering
	\includegraphics[width=\textwidth]{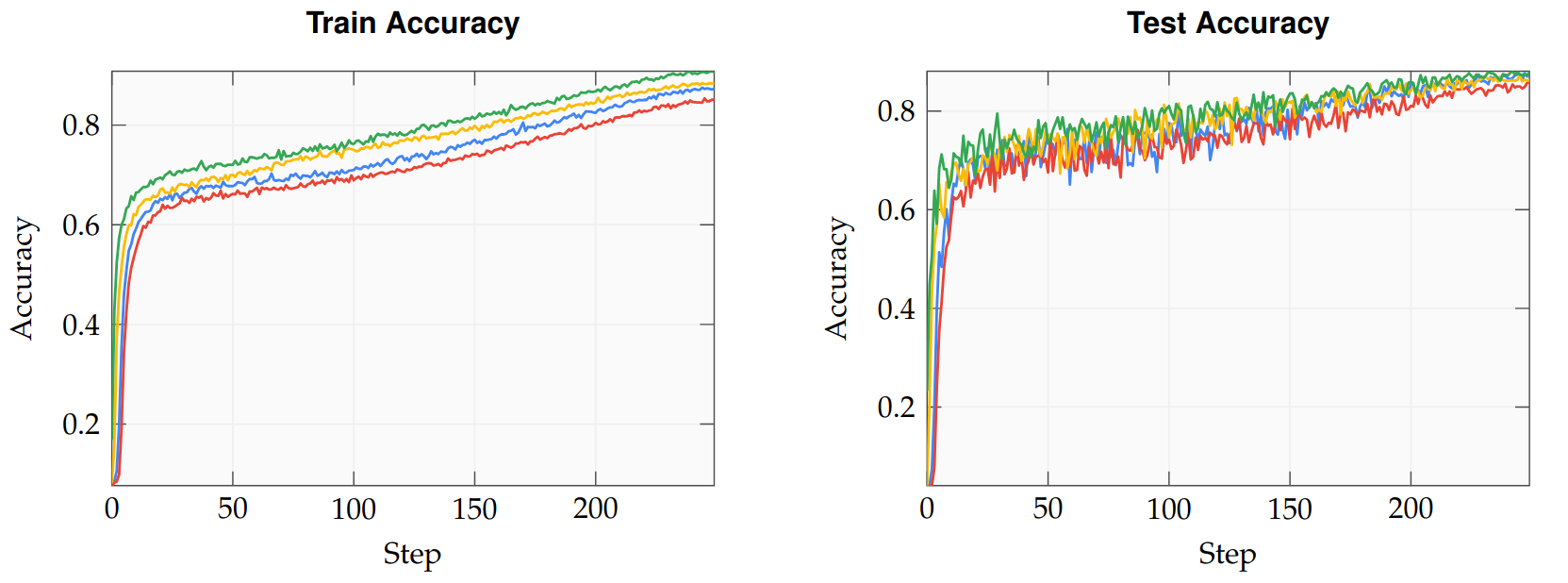}

	\caption{Comparison of Train and Test Accuracy for Varying Degree Values }
	\label{fig:acc2}
\end{figure}

\begin{figure}[htbp]
	
	\centering
	% Subfigure for avg_per_class_accuracy
    \includegraphics[width=\textwidth]{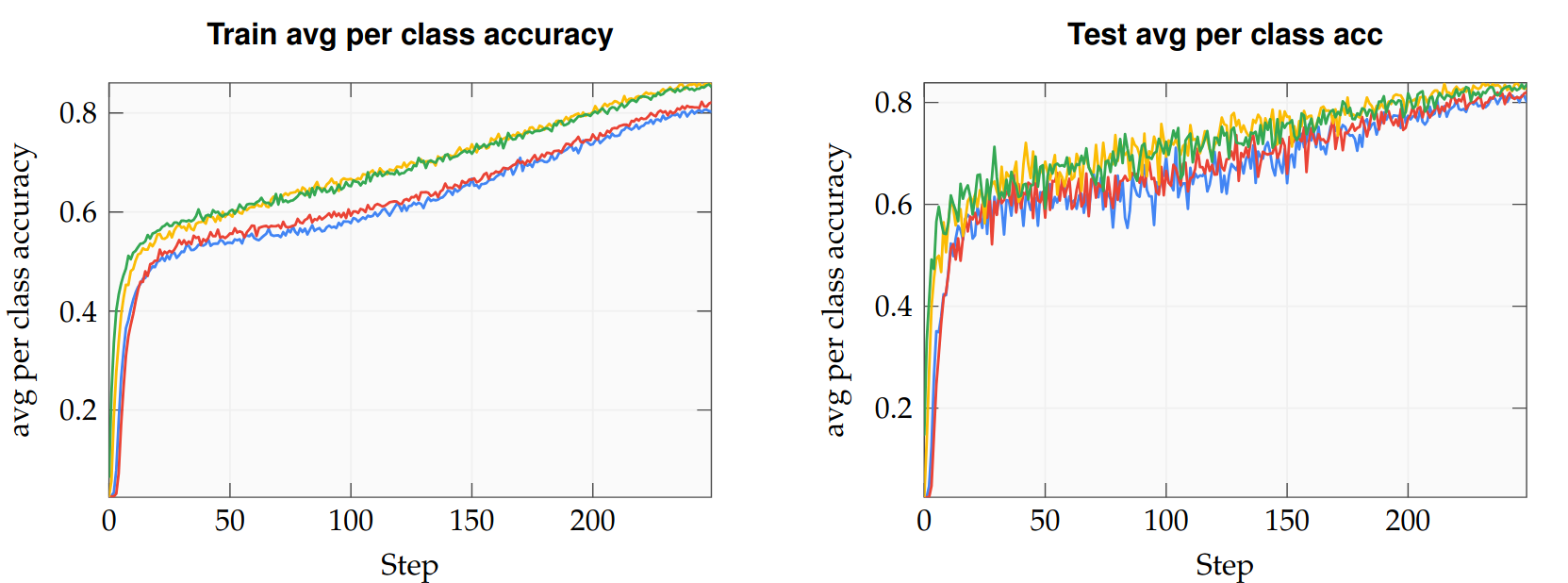}
	\caption{Comparison of Train and Test avg per class accuracy for Varying Degree Values}
	\label{fig:avg_acc2}
\end{figure}

\section{Future Work}
Several aspects warrant further investigation:

\begin{itemize} \item The counterintuitive relationship between polynomial degree and model performance suggests the need for a deeper theoretical analysis of the model's representation capacity.
	\item Future work should extend beyond classification accuracy to examine additional metrics such as inference time and memory efficiency.
	\item The impact of different $\alpha$ and $\beta$ parameters in the Jacobi polynomials remains unexplored and could potentially yield further improvements.
	\item Comparisons when deep layers are stacked should be studied. It would be insightful to analyze how the stacking of layers affects the performance of these models.
	\item Additionally, a comparison should be made when the number of parameters is fixed, as opposed to using just one layer for each model type, to understand the effects of model complexity and depth on performance.
\end{itemize}

\bibliographystyle{unsrt}  

\bibliography{references} % Name of your .bib file without the .bib extension

\end{document}